%% file: root.tex
\documentclass[letterpaper, 10 pt, conference]{ieeeconf}  

\IEEEoverridecommandlockouts                              

\title{\LARGE \bf
UrbanFly: Uncertainty-Aware Planning for Navigation Amongst High-Rises with Monocular Visual-Inertial SLAM Maps
}

\usepackage{amsmath}
\usepackage{mathtools}
\usepackage{booktabs}
\usepackage{float}
\usepackage{xspace}
\usepackage{cite}
\usepackage{graphicx}
\graphicspath{{./figures/}}
\usepackage{caption}
\usepackage{subcaption}
\usepackage[inline]{asymptote}
\usepackage{tikz}
\usepackage{comment}
\usepackage{xcolor}
\usepackage{hyperref}
\usepackage{multirow}
\usepackage{multicol}
\let\labelindent\relax

\usepackage{enumitem}
\usepackage{amsthm}
\usepackage{pgfplots}
\pgfplotsset{width=10cm,compat=1.9}

\usepgfplotslibrary{external}

\setlist[itemize]{leftmargin=*}
\setlist[enumerate]{leftmargin=*}
\usepackage{nomencl}
\makenomenclature
\newcommand{\norm}[1]{\left\lVert#1\right\rVert}

\usepackage[noend]{algpseudocode}
\usepackage[linesnumbered,ruled]{algorithm2e}

\hypersetup{
    colorlinks=true,
    urlcolor=blue
    }
\usepackage{xcolor}
\PassOptionsToPackage{hyphens}{url}\usepackage{hyperref}

\usetikzlibrary{bayesnet}
\usetikzlibrary{arrows, positioning}

\usepackage{array}
\newcommand{\PreserveBackslash}[1]{\let\temp=\\#1\let\\=\temp}

\newcolumntype{C}[1]{>{\PreserveBackslash\centering}p{#1}}
\newcolumntype{L}[1]{>{\PreserveBackslash\raggedright}p{#1}}

\author{Sudarshan S Harithas$^{1}$, Ayyappa Swamy Thatavarthy$^{1}$, Gurkirat Singh$^{1}$, Arun K Singh$^{2}$, K Madhava Krishna$^{1}$
\thanks{$^{1}$are with RRC, IIIT Hyderabad, India
        {\tt mkrishna@iiit.ac.in, \{sudarshan.s, vvsst.ayyappa\}@research.iiit.ac.in, gurkirat.singh@students.iiit.ac.in}}
\thanks{$^{2} $is with University of Tartu, Estonia 
        {\tt\footnotesize aks1812@gmail.com}}
\thanks{$^\dagger$Project page: \href{https://github.com/sudarshan-s-harithas/UrbanFly}{https://github.com/sudarshan-s-harithas/UrbanFly}  } 
}

\begin{document}
\let\@oldmaketitle\@maketitle
\renewcommand{\@maketitle}{\@oldmaketitle
\centering
\vspace{-1mm}

\includegraphics[width=\textwidth]{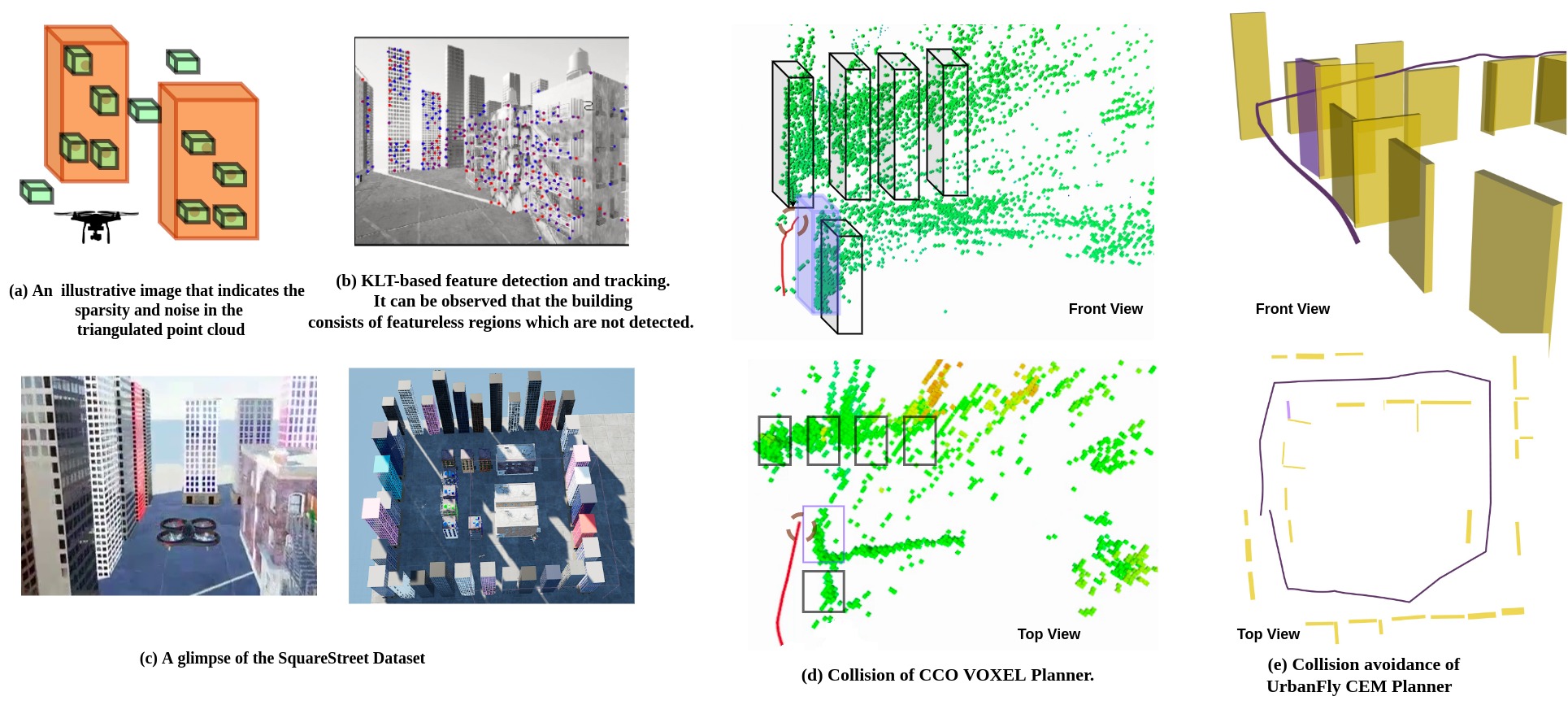}
\vspace{-5.5mm}
\captionof{figure}{\footnotesize{ \textbf{Overview}: Accurate distance measurement to the closest obstacle and associated gradient cannot be expected from a noisy and sparse voxel grid representation built from the triangulated point cloud of a monocular visual-inertial SLAM system (VI-SLAM). This, in turn, poses a critical challenge for trajectory planning. Fig (a) and Fig(b) show that only a few key points that belong to feature-rich regions of the building are detected and tracked by the VI-SLAM resulting in a sparse point cloud (green cuboids in Fig. (a) represent the sparse triangulated point cloud as a discrete voxel). This motivates us to use planes to represent the obstacle instead of the commonly used voxel-grid representation derived from point clouds. Fig. (d)  CCO-VOXEL \cite{ccovoxel}, which plans on voxel-grid representation, fails to plan the trajectory using the triangulated point cloud from VI-SLAM \cite{vins_mono}. CCO-VOXEL uses Octomap as the voxel grid representation, shown as green cubes. The transparent elongated cuboids are the ground truth location of the buildings (obtained from the simulator), and the building in which CCO-VOXEL has collided is highlighted in purple.  Fig. (e) UrbanFly's CEM planner successfully plans a collision-free trajectory while considering the uncertainty-integrated cuboid obstacle representation. Fig. (c) The \textit{SquareStreet} simulation environment that we created to test our planner. Please refer to our $^\dagger$Project page for the video demonstration.   }
}
\label{teaser}
\vspace{-5mm}
}

\maketitle
\thispagestyle{empty}
\pagestyle{empty}

\begin{abstract}

\input{abstract}

\end{abstract}

\vspace{-1.0mm}
\section{INTRODUCTION}
\vspace{-1mm}
\input{Introduction}

\section{Related Work}
\vspace{-1mm}
\label{section:related_work}
\input{Related_work}
\section{Pipeline Overview}
\vspace{-2mm}
\label{section:pipeline}

\input{pipeline}

\section{Monocular Visual-Inertial SLAM based Obstacle Reconstruction} \label{perception_module}
\vspace{-1mm}
\label{section:mono_vins_mapping}

\input{mono_vins_mapping}
\vspace{-1mm}
\section{Uncertainty aware Trajectory Planning} \label{planning_module}
\vspace{-1mm}
\label{section:planning}
\input{planning}

\section{Experiments and Validation}
\vspace{-1mm}
\label{section:results}

\input{results}

\vspace{-1mm}
\section{Conclusions and Future Work}
\vspace{-1mm}
\label{section:conclusion}
\input{conclusion}

\vspace{-2mm}

\bibliographystyle{IEEEtran}
\bibliography{references}

\end{document}

%% file: abstract.tex
We present UrbanFly: an uncertainty-aware real-time planning framework for quadrotor navigation in urban high-rise environments. A core aspect of UrbanFly is its ability to robustly plan directly on the sparse point clouds generated by a Monocular Visual Inertial SLAM (VI-SLAM) backend. It achieves this by using the sparse point clouds to build an uncertainty-integrated cuboid representation of the environment through a data-driven monocular plane segmentation network. Our chosen world model provides faster distance queries than the more common voxel-grid representation. UrbanFly leverages this capability in two different ways leading to two trajectory optimizers. The first optimizer uses gradient-free cross-entropy method to compute trajectories that minimize collision probability and smoothness cost. Our second optimizer is a simplified version of the first and uses a sequential convex programming optimizer initialized  based on probabilistic safety estimates on a set of randomly drawn trajectories. Both our trajectory optimizers are made computationally tractable and independent of the nature of underlying uncertainty by embedding the distribution of collision violations in Reproducing Kernel Hilbert Space. Empowered by the algorithmic innovation, UrbanFly outperforms competing baselines in metrics such as collision rate, trajectory length, etc., on a high fidelity AirSim simulator augmented with synthetic and real-world dataset scenes.

%% file: introduction.tex
With the increasing possibility of Urban Air Mobility (UAM) in the recent future, quadrotors' need to navigate through urban high-rises becomes increasingly inevitable. Monocular vision continues to be a widespread perception modality for quadrotors considering payload, portability, and endurance. Hence navigation based on monocular SLAM maps becomes a natural choice. However, these maps tend to be sparse, noisy, and not represented in the scale of the actual world. We can use the popular  Visual Inertial SLAM (VI-SLAM) \cite{vins_mono} to obtain the quadrotor odometry and consequently bring the maps to metric scale. Yet, uncertainty and sparsity of the point clouds are not entirely alleviated as illustrated in Fig.1 (a)(b) . Thus, it becomes critical to account for this uncertainty during trajectory planning. Unfortunately, existing algorithms for uncertainty-aware planning are not designed to work with systems that rely on monocular vision-based perception. For example, works like \cite{CCO_distance_to_collision}, \cite{Mora_RAL2019}, either assume that the obstacle geometries are precisely known or rely on depth sensors to obtain obstacle geometries during run-time.

In this paper, we present UrbanFly, a novel planning approach that can leverage the capabilities of monocular vision-based perception and at the same time counter the effect of underlying uncertainty. Our approach has two core components: (i) an uncertainty-integrated representation of the environment and (ii) trajectory optimizers specifically designed to exploit the chosen representation. The novelties and benefits of our approach are summarized below.

\begin{figure*}[!t]
    \begin{subfigure}[b]{0.35\linewidth}
    \centering
        \includegraphics[width=\linewidth]{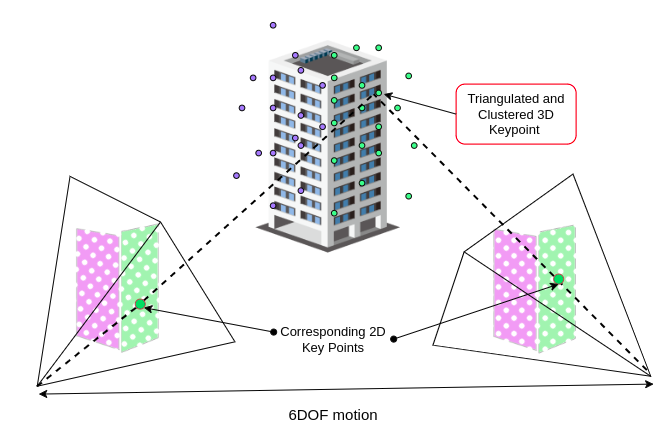}
        \subcaption{ Two subsequent 6DOF camera poses along with the the position obtained from odometry is used to perform triangulation. Furthermore, PlaneRCNN \cite{planercnn} is used for plane segmentation and 2D features and plane masks are used for clustering the 3D triangulated key points.  }
        \label{step1}
    \end{subfigure}
    \hfill
    \begin{subfigure}[b]{0.35\linewidth}
    \centering
        \includegraphics[width=\linewidth]{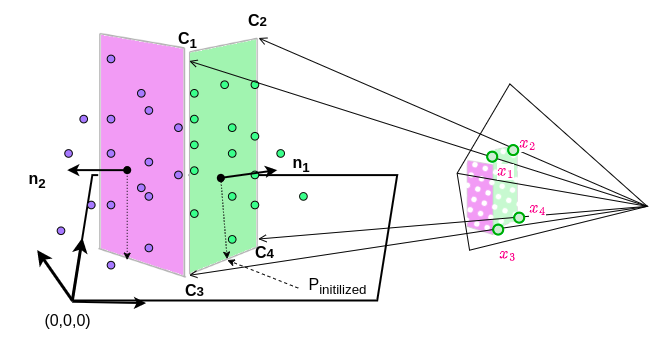}
        \subcaption{ Once the points are triangulated and clustered, we use  RANSAC to obtain the parameters of the Nominal Plane. Note that $\textbf{x}_i$  refers to the corner location of the plane segment on the image plane and $\textbf{c}_i$ is obtained through back-projection ($i \in {1,2,3,4} $  ) from ~\eqref{back_proj}    }
        \label{step2}
    \end{subfigure}
    \hfill
    \begin{subfigure}[b]{0.23\linewidth}
    \centering
        \includegraphics[width=0.45\textwidth, height=0.35\textwidth]{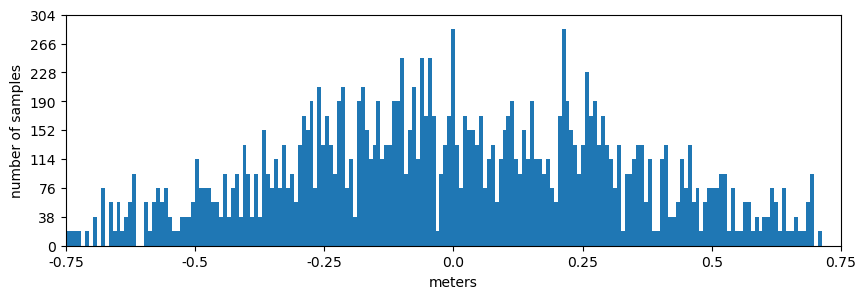}
        \includegraphics[width=0.45\textwidth, height=0.35\textwidth]{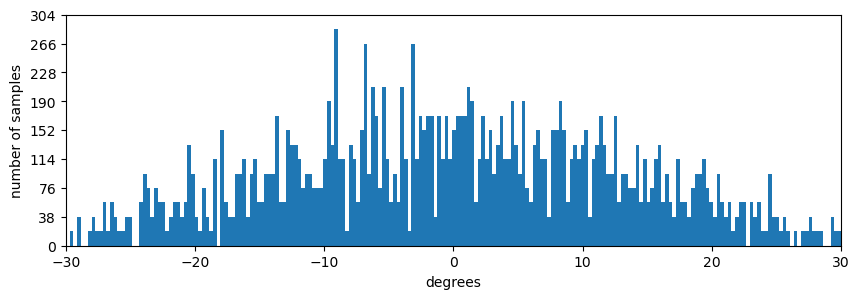}\\[1mm]
        \includegraphics[width=0.45\textwidth, height=0.35\textwidth]{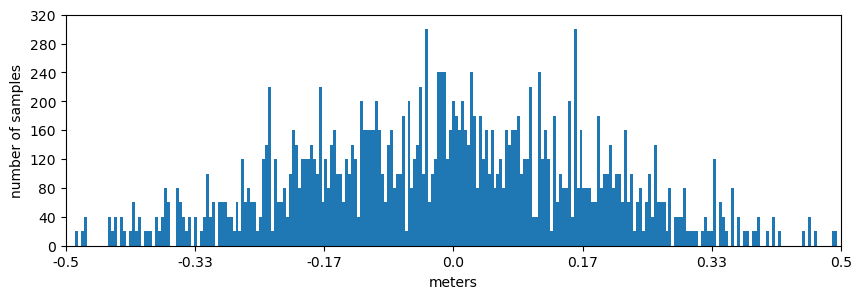}
        \includegraphics[width=0.45\textwidth, height=0.35\textwidth]{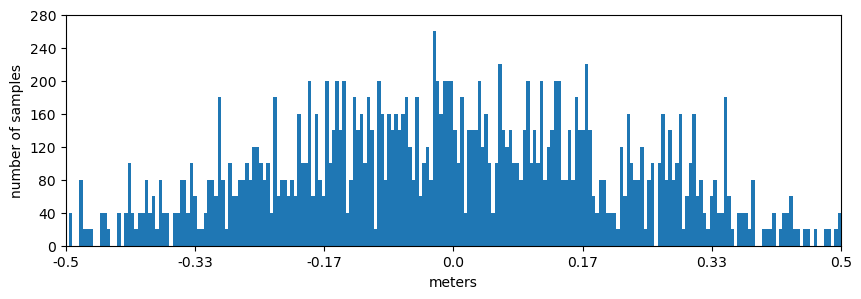}
        \subcaption{ Error Histogram for the plane parameters, the yaw error is shown in top-left, the origin error is shown in the top right, and the bottom row refers to the error in size length and width, respectively.   }
        \label{histogram}
    \end{subfigure}

    \caption{The figure illustrates the procedure for obtaining the nominal plane estimates ( description in Section \ref{nominal_plane}, Fig. (a) represents the first step which includes triangulation, segmentation, and clustering, Fig. (b) represents the second setp of RANSAC, Fig. (c) The error histograms related to various plane parameters.         } 

\end{figure*}

\noindent \textbf{Algorithmic Contribution:}  UrbanFly uses a data-driven neural network to segment the high-rises in the image into their planar constituents. The resulting plane masks along with the point cloud obtained from the triangulation from monocular VI-SLAM are used to build a cuboid representation of the obstacles (skyscrapers).  We show how the non-parametric point cloud uncertainty can be mapped to the uncertainty in the parameters of the cuboid and finally to the distribution over distance to the closest obstacle. The sample estimates of closest obstacle distance can be coupled with the notion of distribution embedding in Reproducing Kernel Hilbert Space (RKHS) to estimate collision probability. We propose two trajectory optimizers capable of minimizing this estimate along with the conventional smoothness cost. Our first optimizer uses the gradient-free cross-entropy method to compute trajectories that explicitly trade-off collision risk and smoothness cost. Our second optimizer is a simplified but computationally faster version of the first. It computes trajectories that avoids inflated version of the cuboid obstacles using a sequential convex programming (SCP) based optimizer. The unqiue aspect of our SCP is that it is initialized based on probabilistic safety estimates on a set of randomly drawn trajectories.




\noindent \textbf{State-of-the-Art Performance:}
\begin{itemize}
    \item  Our work is one amongst the few rare papers to demonstrate robust and collision-free autonomous flight with a single camera system.
    
    \item We outperform recent work \cite{ccovoxel} in success-rate, smoothness cost and traversal length. This is because \cite{ccovoxel} struggles to reliably build the voxel-grid representation of the world based on the sparse point clouds generated by the VI-SLAM.
    
\end{itemize}

\vspace{-1.5mm}

%% file: Related_work.tex
\noindent \textbf{Monocular SLAM based Planners:} Literature dealing with trajectory planning and/or navigation with monocular SLAM is sparse when compared with a much larger volume of literature that deals with planning with SLAM \cite{dissa}. A large part of planning with SLAM focuses on actively moving to places where the state uncertainty (often measured as the trace of the state co-variance) is low \cite{dissa}, \cite{dissa_slam_plan}. In contrast, the Monocular SLAM frameworks tend to rely on heuristics \cite{mostegel} due to the difficulty of estimating accurate co-variance in a monocular setting. In \cite{scaramuzza} photometric co-variance is considered with state co-variance propagated over the tangent manifold. Nonetheless, this technique relies on Gaussian noise models, which we relax in our current work.
Our approach also goes beyond \cite{ocpncy_grid_margarita} that shows navigation with a camera but does not consider the underlying uncertainty during planning.

\noindent \textbf{Risk Aware Planning:} Chance constrained optimization (CCO) \cite{chance_contrained} has proven to be a popular paradigm for trajectory planning under uncertainty, wherein the probability of collision avoidance is balanced with trajectory smoothness, arc-length, etc. Most existing works, such as \cite{Mora_RAL2019}, rely on Gaussian uncertainty model and pre-defined geometry in obstacle shapes. In contrast, sampling-based approaches such as \cite{blackmore_chance_constrained} can adapt to more general settings in uncertainty and obstacle geometry but are computationally prohibitive.

Only recently, the notion of CCO was extended to settings where the world is described through a voxel-grid representation of the world derived from point clouds \cite{ccovoxel}. Our current works extends this line of work further to a more challenging setting where the point clouds generated by the perception system are very sparse and noisy.



\vspace{-2mm}


%% file: pipeline.tex
\noindent \textbf{Perception Module:} Our perception module (detailed in section \ref{perception_module}) satisfies a two-fold objective:

\begin{enumerate}
    \item \textbf{State estimation:} An on-board monocular VI-SLAM (VINS-Mono) \cite{vins_mono} is used to perform state estimation. The state includes the 6DOF pose of the quadrotor and an upto scale 3D key point locations.
    \item \textbf{Uncertainty aware Mapping}: The sparse and noisy point clouds from monocular VI-SLAM are used to represent the obstacles in the environment as axis-aligned cuboids. The perception module also quantifies the uncertainty in cuboid parameters. 
\end{enumerate}

\noindent \textbf{Planning Module:} The planning module has a single objective: to perform uncertainty aware trajectory planning by leveraging upon the cuboid representation of the world obtained from the perception module. We develop two trajectory optimizers that are capable of accomplishing this objective. Our first optimizer follows the stochastic chance-constrained optimization paradigm and formulates a trade-off between probabilistic safety and smoothness cost. We use the gradient-free CEM to solve the resulting optimization problem. Our second optimizer, called SCP-MMD can be considered as a simplified but computationally faster version of the first. We draw several random trajectories and compute the collision-avoidance probability estimates along each of these to compute the safest trajectory. We then use sequential convex programming with collision avoidance constraints induced by cuboid planes to further refine the safest trajectory.  The formulation of our trajectory optimizer has been detailed in section \ref{planning_module}.

%% file: mono_vins_mapping.tex
\subsection{Monocular visual-inertial SLAM}
\label{subsection:monocular_vins}
\noindent Although a detailed description of monocular visual-inertial SLAM \cite{vins_mono} is beyond the scope of this paper, we briefly describe the steps involved in estimating the odometry from images and IMU.

The system starts with measurement pre-processing, wherein 2D key points are extracted and tracked. Simultaneously, we obtain the relative motion constraints from the IMU through a process popularly known as IMU pre-integration in SLAM literature \cite{imu_pre_int}. Then the initialization procedure provides all necessary values, including pose, velocity, gravity vector, gyroscope bias, and three-dimensional (3D) key points (the point clouds), for bootstrapping the subsequent sliding-window-based nonlinear optimization. Finally, the drift is minimized through a global pose graph optimization. We use the resulting camera pose to transfer the 3D key points into the world frame that are subsequently used in step \ref{subsection:plane_init} for plane reconstruction.

\subsubsection*{\textbf{Assumption}} We assume that the buildings are planar and feature-rich. We expect a reliable state estimate from the SLAM but acknowledge the uncertainty in the estimated 3D planes obtained from the sparse triangulated point clouds of the monocular visual-inertial SLAM. We also assume that the VI-SLAM estimate of the gravity axis is sufficiently accurate. 
\vspace{-1mm}

\vspace{-1mm}

\subsection{3D plane reconstruction}
\label{subsection:plane_init}
\noindent \textbf{Notations:} In the section \ref{nominal_plane} \ref{analysis_error} and \ref{sample_multiple_planes}, we use the symbols $\mathbf{n} , \mathbf{p}, \mathbf{s}$ to refer to the normal, 2D plane origin and size of the plane respectively. However, the nominal plane estimates are denoted by $\mathbf{n_0}, \mathbf{p_0}, \mathbf{s_0}$. Furthermore, in section \ref{sample_multiple_planes} where we sample multiple planes, the parameters of the individual samples are represented by the symbols $\mathbf{n_i} , \mathbf{p_i}, \mathbf{s_i}$ (i $\in$ {1,2, .., N} ). 
 
\subsubsection{Nominal Plane parameter estimation } \label{nominal_plane}

Estimating the parameters of the nominal plane involves two steps; in the first step, we perform plane segmentation, triangulation, and clustering. In the second step, we complete the process by estimating the plane parameters using RANSAC.

\begin{enumerate}[label=(\roman*)]
    \item \textbf{Plane segmentation, triangulation, and clustering}:  We use PlaneRCNN \cite{planercnn}, a data-driven network, to generate the 2D plane segment masks from the RGB image. Each 3D key point triangulated by VI-SLAM is then associated with a plane if the corresponding 2D key point observation falls within a plane segment mask in the image. This results in the clustering of 3D points. An illustrative diagram of this step is shown in Fig. ~\ref{step1} 
    
    \item \textbf{Parameter estimation  of the nominal plane:} The finite nominal plane is parameterized through three quantities namely: the plane normal ($\mathbf{n_0} = [n_{0x},n_{0y},n_{0z}]$), 2D-plane origin ($ \mathbf{p_0} \in R^{2}$)  obtained by projecting the center of the building plane onto the ground plane, and the length and height of the plane which is referred to as the size ($\mathbf{s} = [l,w] $) (where $l$ is length and $w$ is width of the plane). For each cluster obtained from the previous step, we perform RANSAC to estimate the nominal normal $(\mathbf{n_0})$, which leads to a parametric equation of the plane $n_{0x}x+n_{0y}y+n_{0z}z +d_{p}=0 $, the parameter $d_{p}$ is a scalar constant. 

    The parameters obtained through RANSAC are for an infinite plane, whereas the buildings are finite. The finite size ($\mathbf{s}$) and the 2D ground plane origin ($\mathbf{p_0}$) of the nominal plane are determined by estimating the corners of the building. The corners are obtained by back-projecting the corners in the plane segment mask  using  \eqref{back_proj}. The vector $\textbf{c}_{i}$ (i $\in$ {1,2,3,4}) is the $i^{th}$ corner, $\mathbf{\Tilde{K}}$ is the camera matrix (that includes both the intrinsics and extrinsic) and $\mathbf{\Tilde{x_i}}$ is the pixel location of the $i^{th}$ plane corner ( obtained from the plane mask) in homogeneous form. The depth value $d_i$ is given by the equation  $d_i = | \frac{d_{p}}{ n_0.(\mathbf{\Tilde{K}^{-1}} \mathbf{\Tilde{x_i}} ) } | $. The process of using RANSAC for plane parameter estimation and back-projection has been illustrated in Fig. ~\ref{step2}.
    
    \begin{equation} \label{back_proj}
        \textbf{c}_{i} = d_{i}\mathbf{\Tilde{K}^{-1}}\mathbf{\Tilde{x_i}}
    \end{equation}

\end{enumerate}

In order to make the plane segmentation process faster and effective, we propagate the detected plane segment mask to next four frames using dense optical flow as described in \cite{farneback}, i.e. PlaneRCNN is used for every 1 out 5 RGB image frames. The obtained plane masks are used for clustering the 3D key points in the following section.


\subsubsection{Modelling and Analysis the of SE(2) Uncertainty in 3D planes } \label{analysis_error}
\noindent The triangulated point cloud from the monocular VI-SLAM is highly noisy and can lead to inaccurate detection of planes. We map the uncertainty present in the triangulated point cloud to the uncertainty in the parameters of the estimated plane. To this end, we assume that building facades in the scene are vertical. This results in only SE(2) (a 3DOF) movement for the building plane viz. 2D planar translation along the ground plane and a 1D yaw rotation along the axis perpendicular to the ground plane.  

As explained previously, the plane is modeled through three parameters, the normal ($\mathbf{n}$), plane origin $\mathbf{p}$, and size $\mathbf{s}$.  To model the error, we need access to the ground truth parameters of the plane; we set up an experiment in Unreal  \cite{unrealengine}, a high-fidelity gaming engine, to obtain the ground truth parameters of the plane through simulation. To determine the error, a human pilot would fly the quadrotor through the simulation environment (the environment details are given in section \ref{Simulation_env_Details}), and the procedure mentioned in the previous section would be performed to obtain the nominal plane estimates. The difference between the observed (the nominal values) and ground truth measurements would be used to model the error. The uncertainty in each of the plane parameters is modeled as follows:

\begin{enumerate}[label=(\roman*)]
    \item Normal $\mathbf{n}$: The normal determines the orientation of the plane, and from our SE(2) assumption, we notice that this orientation is the 1D yaw (viz., the rotation along the axis perpendicular to the ground plane). Therefore, we parameterize the uncertainty in the normal through the error in the yaw (orientation) denoted by $\Delta_{\psi}$. Through simulation, we obtain the ground truth orientation of the plane denoted by $\psi_{GT}$, and we further obtain the nominal plane estimates; we measure the observed orientation given by $\psi_{0}$. Where $\psi_{0} = \arctan{ \frac{n_{0y}}{n_{0x}} }$ for the observed plane normal $ \mathbf{n_0}$. and $\Delta_{\psi} = \psi_{0} - \psi_{GT}$. From Fig. ~\ref{histogram} (top left)  we observe that this error is non-parametric.  
    \item Size $\mathbf{s}$: The size of the plane refers to the length and width of the plane and it is measured in meters. Through the simulator, we obtain the ground truth size of the plane, and the nominal plane parameters are obtained as explained above. The difference between them would result in the error histogram shown in Fig. ~\ref{histogram} (bottom row). We further observe that the size distribution does not follow the parametric Gaussian form. 
    \item Plane origin $\mathbf{p}$: The error in the plane origin is represented by a vector in $R^{2}$.  It is obtained by taking the difference between the true plane origin (obtained from simulation) and the observed plane origin. The distribution of plane origin is also non-Gaussian, as shown in Fig. ~\ref{histogram} (top right).
\end{enumerate}

\subsubsection{Sampling multiple planes} \label{sample_multiple_planes}

Assume that we can fit a non-parametric distribution to the plane parameters and sample $\Delta_{\psi_i} , \Delta_{s_{i}}$ and $\Delta_{p_i}$ from that. Note the fitted form need not be even analytically known. We can generate multiple samples of the plane parameters through \eqref{error_type}. The process of sampling multiple planes by perturbing individual plane parameters and combining them is illustrated in Fig. ~\ref{fig:uncertainity_flow}


 \vspace{-2mm}
\begin{equation}\label{error_type}
      \psi_i = \psi_{0} + \Delta_{\psi_i},
        \mathbf{s_{i}} = \mathbf{s_{0}} + \Delta_{s_i}, 
        \mathbf{p_{i}} = \mathbf{p_{0}} + \Delta_{p_i}
\vspace{-2mm}
\end{equation}

\subsubsection{Cuboid approximation of the 3D plane segment}
\label{subsection:cuboid_approx}

The primary objective of modeling the uncertainty in the plane is to exploit its properties in a downstream task such as trajectory planning. Performing collision checks is of key importance in trajectory planning and can be rapidly performed by querying an analytical \textit{Signed Distance Function} (SDF). To take advantage of the geometrical properties of the SDF, we consider each finite plane as a cuboid of infinitesimal thickness.

%% file: planning.tex
\subsection{Distance Queries to Cuboid}
\label{subsection:distance_queries}



\noindent The distance to the surface can be obtained by querying the analytical SDF as given in ~\eqref{box_sdf}; this allows for rapid collision checks.  For a given query point $\mathbf{q}= [x,y,z] $ (measured in world frame), the resulting distance $d_{ijk}$ represents the distance measured to the cube using the $i^{th}$ sample from the yaw angle distribution $\psi_{i}$, the $j^{th}$ sample from the size distribution $\mathbf{s_{j}}$  and $k^{th}$ sample from the origin distribution $\mathbf{p_{k}}$, the matrix $\mathbf{T_{i}}$ is a transformation matrix defined by  ~\eqref{transform1} that transforms a given query point from the world frame to the local frame of the cuboid, $|.|$ denotes the absolute value function. The $\max$ function would perform the element-wise maximum operation and would return a zero vector if the query point is on or within the cuboid. The $\norm{ .. }_{2}$ would compute the $l_{2}$ norm of the resulting vector. 

\begin{equation}\label{box_sdf}
    d_{ijk} = \norm{ \max \Big(   \big(  \lvert \mathbf{T(\psi_{i} , \mathbf{p_{k}} ) } \mathbf{q} \rvert - \mathbf{s_{j}} \big) , \mathbf{0}_{3 \times 1}  \Big)}_{2}
\end{equation}

 

\begin{figure}[!t]
    \centering
\includegraphics[width=\linewidth]{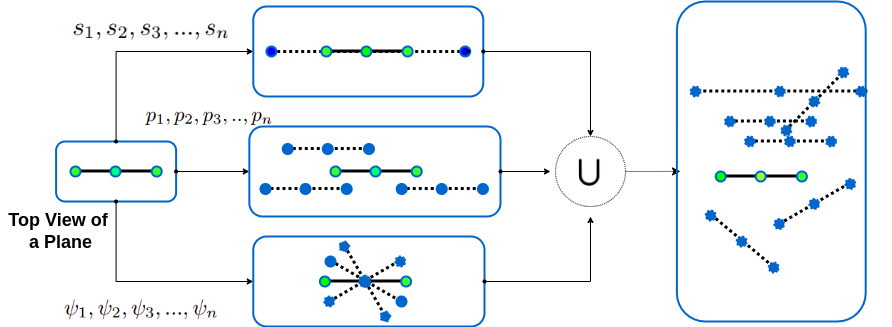}
    \caption{ In the top view, a plane appears as a line, as shown in the extreme left (the thick line with three green dots indicates the nominal plane estimate). Perturbing each plane parameters individually can distort the plane in different ways, as shown through the middle blocks. The symbol $\cup$ represents an union of all the random perturbations that results in a total of $n_\psi \times n_s \times n_o$ samples of the plane; an illustration of the sample set is shown on the extreme right. }
    \label{fig:uncertainity_flow}
    \vspace{-7mm}
\end{figure}

\noindent\begin{minipage}{.5\linewidth}\footnotesize 
\begin{equation*}  
  \mathbf{T_{i} } =  \left[  \begin{array}{c|c}
\mathbf{R_{i}^{T}} & -\mathbf{R_{i}^{T}}\mathbf{p}_k \\
\hline
\mathbf{0}_{1 \times 3 } & 1
\end{array}\right]  
\end{equation*}
\end{minipage}%
\begin{minipage}{.5\linewidth} \footnotesize
\vspace{-3mm}
\begin{equation} \label{transform1}
 \mathbf{R_{i}^{T}} =  \begin{bmatrix}
\cos \psi_{i} & -\sin \psi_{i} & 0\\
\sin \psi_{i} & \cos \psi_{i} & 0\\
0 & 0 & 1 \\ 
\end{bmatrix}
\end{equation}
\end{minipage}

 Since we draw  $\mathit{n_{\psi}}$, $\mathit{n_{s}}$ and  $\mathit{n_{o}}$  samples from the black-box error distribution of the yaw, size and origin uncertainties, we get a total of $\mathit{n_{\psi}}\times \mathit{n_{s}} \times \mathit{n_{o}} $ distance measurements for each query point. The process of relating distance queries as collision constrains is described in the next section \ref{subsection:constraint_violation} .

\vspace{-2mm}
\subsection{Surrogate for Collision Probability}
\label{subsection:constraint_violation}
\noindent \textbf{Distribution over Constraint Violations:} Our vision-based perception module enforces some additional requirements on the planning side. It is not merely enough to maintain a large conservative distance from the obstacles (sky-scrapers) as in that case, we lose the coarse features that are present at the boundaries of the obstacles. On the other hand, flying very close to the obstacles may compromise safety. We thus propose to bound the distance estimates $d_{ijk}$ between $r_{min}$ and $r_{max}$. To this end, we formulate the following constraint violation function. 
\begin{equation} \label{const_violation}
    \overline{f}_{ijk} = \max ( d_{ijk} - r_{max} , 0) + \max( r_{min} - d_{ijk} , 0)  
\end{equation}
\vspace{-6mm}




Now, as explained in Section \ref{subsection:distance_queries}, $d_{ijk}$ are in fact sample approximations of some unknown true estimate of distance $\hat{d}$ to the closest obstacle. Thus, \eqref{const_violation} maps the distribution over distance estimates to distribution of constraint violation. Moreover, $\overline{f}_{ijk}$ are the sample approximations of the true distribution of $\overline{f}$ which we henceforth call as $p_{\overline{f}}$. Although computing the exact analytical shape of the distribution $p_{\overline{f}}$ is intractable we can make the following remarks about its shape.


\newtheorem{remark}{Remark}
\begin{remark} \label{des_dirac}
The best possible shape of $p_{\overline{f}}$ is given by a Dirac Delta distribution $p_{\delta}$ . 
\end{remark}
\begin{remark}\label{dirac_sim}
As $p_{\overline{f}}$ becomes increasingly similar to $p_{\delta}$, the probability that the quadrotor is at-least $r_{min}$ away from the obstacle an at-most $r_{max}$ away from the obstacle increases.
\end{remark}

\noindent \textbf{Probabilistic Safety :} Remarks \ref{des_dirac}, \ref{dirac_sim} forms the basis of our notion of probabilistic safety. Instead of directly measuring the probability of collision avoidance; we define the following surrogate.

\newtheorem{definition}{Definition}
\begin{definition}
Let $l_{dist}(p_{\overline{f}}, p_{\delta} )$ be a positive semi-definite function that quantifies the similarity between the $p_{\overline{f}}$, and $ p_{\delta}$. That is, $l_{dist}$ decreases as the distribution becomes similar. Then, $l_{dist}$ can act as an estimate of the probability of collision avoidance.
\end{definition}




\subsection{Maximum Mean Discrepancy as $l_{dist}$} 
\label{subsection:MMD}
\vspace{-1mm}

\noindent The Maximum Mean Discrepancy or MMD is a popular approach to quantify the similarity between two distributions in \textit{Reproducing Kernel Hilbert Space} (RKHS) using just the sample level information. Thus, it can be a potential option for $l_{dist}$. To this end, let $\mu_{p_{\overline{f}}}$ and $\mu_{p_{\delta}}$ represent the RKHS embedding of $p_{\overline{f}}$ and $p_{\delta}$ computed in the following manner.

\vspace{-3mm}
\begin{equation}
 \footnotesize 
\mu_{p_{\overline{f} }} = \sum_{i,j, k=1}^{n_{\psi}, n_{s}, n_o } \alpha_i \beta_j \gamma_{k} k(\overline{f }_{ijk}  , \cdot ) \label{kernel_mean_fover1} 
\end{equation}
\begin{equation}
\mu_{p_{\delta}} =\sum_{i,j,k=1}^{n_{\psi}, n_{s}, n_o } \lambda_i \varphi_j \theta_{k} k( 0 , \cdot ) \label{kernel_mean_dirac}
\end{equation}


\noindent Where $k(.,.)$ is the kernel function (we use an \textit{Radial Basis Function} (RBF) kernel) , $\overline{f }_{ijk}$ represents the constraint violation function defined in \eqref{const_violation}. The constants $\alpha_{i}$, $\beta_{j}$ and $\gamma_{k}$ are the weights associated with the $\overline{f }_{ijk}$ sample of the constraint violation function. Similarly, $\lambda_{i}$, $\varphi_{j}$ and $\theta_{k}$ are the weights associated with the sample of the Dirac-Delta distribution. Typically, these weights can be just set to $\frac{1}{n}$. Alternately, we can adopt a more sophisticated approach based on a reduced set or Auto-Encoder-based dimensionality reduction \cite{ccovoxel}. Please note that \eqref{kernel_mean_dirac} follows from the fact that the samples from a Dirac-Delta distribution are all zeros. Using ~\eqref{kernel_mean_fover1} and ~\ref{kernel_mean_dirac} we will proceed to derive the following algebraic form for ($\mathit{l_{dist}}$) as given in ~\eqref{mmd_ldist}.  \vspace{-3mm}
\begin{equation}
\small    l_{dist} = \left\Vert \mu_{p_{\overline{f}}} - \mu_{p_{\delta}} \right \Vert_{2}^{2}
    \label{mmd_ldist}
\vspace{-1mm}
\end{equation}

It is worth noting that MMD is a function of the $d_{ijk}$, which in turn is a function of the query point. In the context of planning, the query points will belong to the trajectory. 



\subsection{Matrix Form of MMD}
\label{subsection:computing_MMD}
\noindent We can efficiently evaluate $l_{dist}$ through the so-called kernel trick. To show how, we expand \eqref{mmd_ldist} as follows



\vspace{-3mm}
\begin{align} 
\footnotesize
    \lVert \mu_{p_{\overline{f}}} -  \mu_{p_{\delta}} \lVert ^{2} = & \langle \mu_{p_{\overline{f}}} , {\mu_{p_{\overline{f}}}} \rangle - 2 \langle {\mu_{p_{\overline{f}} }}, {\mu_{p_{\delta}}} \rangle +   \langle \mu_{p_{\delta}} , \mu_{p_{\delta}} \rangle
\label{MMD_expand}
\vspace{-2mm}
\end{align}

Substituting the kernel mean functions \ref{kernel_mean_fover1} and \ref{kernel_mean_dirac} in \ref{MMD_expand}  

\begin{subequations} \label{kernel_trick}
\footnotesize
\begin{align}
    &\langle \mu_{p_{\overline{f}}}, \mu_{p_{\overline{f}}} \rangle = 
    \langle \sum_{i,j,k=1}^{n_{\psi}, n_{s}, n_o }  \alpha_i \beta_j \gamma_k k(\overline{f}_{ijk}  , \cdot ) , \sum_{i,j,k=1}^{n_{\psi}, n_{s}, n_o } \alpha_i \beta_j \gamma_{k} k(\overline{f}_{ijk}  , \cdot ) \rangle \\
&\langle \mu_{p_{\overline{f}}}, \mu_{p_{\delta}} \rangle = \langle \sum_{i,j,k=1}^{n_{\psi}, n_s, n_o} \alpha_i \beta_j \gamma_{k} k(\overline{f}_{ijk}  , \cdot ), \sum_{i,j,k=1}^{n_{\psi}, n_s, n_o} \lambda_p \varphi_j \theta_{k} k (0, \cdot) \rangle \\ 
&\langle \mu_{p_{\delta}}, \mu_{p_{\delta}} \rangle = 
\langle \sum_{i,j,k=1}^{n_{\psi},n_s,n_o } \lambda_{i} \varphi_{j} \theta_k k (0, \cdot), \sum_{i,j,k=1}^{n_{\psi},n_s,n_o } \lambda_i \varphi_j \theta_k k (0, \cdot) \rangle
\end{align}
\end{subequations}

\noindent Using the kernel trick, \eqref{kernel_trick} can be expressed in matrix form as given below 
\vspace{-7mm}

\begin{align}
    \lVert \mu_{p_{\overline{f}}} - \mu_{p_{\delta}} \lVert ^2 = &
 \sum_{k =1}^{n_{o}}  \mathbf{C_{\alpha \beta / \gamma_{k} } } \mathbf{K}_{\overline{f}\overline{f}}^{k}  \mathbf{C_{\alpha \beta/ \gamma_{k} }^T} - \nonumber \\  2\mathbf{C_{\alpha\beta /\gamma_{k} }} \mathbf{K}_{\overline{f}\delta}^{k} \mathbf{C_{\lambda\varphi / \theta_{k} }^T} + & \mathbf{C_{\lambda\varphi / \theta_{k} }} \mathbf{K}_{\delta\delta}^{k}   \mathbf{C_{\lambda\varphi /\theta_{k}}^T} 
\end{align}

where $ \mathbf{C_{\alpha \beta/ \gamma_{k} }}$ and $ \mathbf{C_{\lambda\varphi/ \theta_{k} }}$ are the weight vectors 
\vspace{-5mm}

\begin{gather*}
 C_{\alpha\beta / \gamma_k } = \begin{bmatrix}
\alpha_{0}\beta_{0}\gamma_k , \alpha_{0}\beta_{1}\gamma_k , ... , \alpha_{n}\beta_{n}\gamma_k
\end{bmatrix}; \\
C_{\lambda\varphi / \theta_k } = \begin{bmatrix}
\lambda_{0}\varphi_{0}\theta_k , \lambda_{0}\varphi_{1}\theta_k, ... , \lambda_{n}\varphi_{n}\theta_k
\end{bmatrix} 
\vspace{-2mm}
\end{gather*}
\vspace{-5mm}


\noindent\begin{minipage}{.5\linewidth}\scriptsize
\begin{align*}
 \scriptsize 
\mathbf{K}_{\overline{f}\overline{f}}^{k} = \begin{bmatrix}
\mathbf{K}_{00}^{k}  & \mathbf{K}_{01}^{k} & \dots & \mathbf{K}_{0n}^{k}  \\
\mathbf{K}_{10}^{k} & \mathbf{K}_{11}^{k} & \dots & \mathbf{K}_{1n}^{k}  \\
\vdots & \vdots & \ddots & \vdots \\
\mathbf{K}_{n0}^{k} & \mathbf{K}_{n1}^{k} & \dots & \mathbf{K}_{nn}^{k}  \\
\end{bmatrix} 
\end{align*}
\label{kernel_mean_fover}
\end{minipage}%
\begin{minipage}{.5\linewidth}\scriptsize
\vspace{-3mm}
\begin{align*} \scriptsize
\mathbf{K}_{\overline{f}\delta}^{k} =  \begin{bmatrix}
\mathbf{K}_{\delta 00}^{k} & \mathbf{K}_{\delta 01}^{k} & \dots & \mathbf{K}_{\delta 0n}^{k}  \\
\mathbf{K}_{\delta 10}^{k} & \mathbf{K}_{\delta 11}^{k} & \dots & \mathbf{K}_{\delta 1n}^{k}  \\
\vdots & \vdots & \ddots & \vdots \\
\mathbf{K}_{\delta n0}^{k} & \mathbf{K}_{\delta n1}^{k} & \dots & \mathbf{K}_{\delta nn}^{k}  \\
\end{bmatrix}
\end{align*}
\end{minipage}

\begin{align*}\footnotesize
\mathbf{K}_{ab}^{k}  = 
{\begin{bmatrix}
k(\overline{f}_{a0k} ,\overline{f}_{b0k} ) & \dots & k(\overline{f}_{ank}, \overline{f}_{b0k} ) \\
k(\overline{f}_{a0k},\overline{f}_{b1k})  & \dots & k(\overline{f}_{ank}, \overline{f}_{b1k}) \\
\vdots & \vdots & \ddots & \vdots \\
k(\overline{f}_{a0k},\overline{f}_{bnk})  & \dots & k(\overline{f}_{ank}, \overline{f}_{bnk}) \\
\end{bmatrix}}
\end{align*}


\begin{align*}\footnotesize
\mathbf{K}_{\delta ab}^{k} = \begin{bmatrix}
k(\overline{f}_{a0k},0) & k(\overline{f}_{a1k}, 0) & \dots & k(\overline{f}_{ank}, 0) \\
k(\overline{f}_{a0k},0) & k(\overline{f}_{a1k}, 0) & \dots & k(\overline{f}_{ank}, 0) \\
\vdots & \vdots & \ddots & \vdots \\
k(\overline{f}_{a0k},0) & k(\overline{f}_{a1k}, 0) & \dots & k(\overline{f}_{ank}, 0) \\
\end{bmatrix}
\vspace{-2mm}
\end{align*}
\[ \mathbf{K}_{\delta\delta}^{k}  = \mathbf{1}_{n_{\psi}n_{s}  \times n_{\psi}n_{s} } \vspace{-2mm}\]

Furthermore, \cite{ccovoxel} proposes an effective method to utilize supervised learning to map distance measurements to a low dimension feature space, this was found to increase the computational speed significantly.  Here we adopt a similar approach for the computation of MMD.  
\vspace{-3mm}
\subsection{CEM Planner} \label{subsection:cem_planner}

\subsubsection{\textbf{Trajectory Parameterization}}
The \textit{CEM Planner} uses a polynomial parameterization of the trajectory, defined as  
\vspace{-1mm}
\begin{equation}
\small
\begin{bmatrix}
x(t_1)\\
x(t_2)\\
\dots\\
x(t_n)
\end{bmatrix} = \textbf{P}\textbf{c}_{x}, \begin{bmatrix}
\dot{x}(t_1)\\
\dot{x}(t_2)\\
\dots\\
\dot{x}(t_n)
\end{bmatrix} = \dot{\textbf{P}}\textbf{c}_{x}, \begin{bmatrix}
\ddot{x}(t_1)\\
\ddot{x}(t_2)\\
\dots\\
\ddot{x}(t_n)
\end{bmatrix} = \ddot{\textbf{P}}\textbf{c}_{x}.
\label{param}
\vspace{-1mm} 
\end{equation}
\noindent where, $\textbf{P}, \dot{\textbf{P}}, \ddot{\textbf{P}}$ are matrices formed with time-dependent basis functions ($x(t_{i})$) (e.g polynomials) and $\textbf{c}_{x}$ are the coefficients associated with the basis functions. Similar expressions can be written for $y(t), z(t)$  in terms of coefficients $\textbf{c}_y, \textbf{c}_z$, respectively.

\subsubsection{\textbf{Optimization Problem}}

\noindent The CEM-based planner minimizes the following cost function, for a given weight $w$. \vspace{-5mm}

\begin{equation}
    \min_{\textbf{c}_x, \textbf{c}_y, \textbf{c}_z} l(\textbf{c}_x, \textbf{c}_y, \textbf{c}_z)+w l_{dist}(p_{\overline{f}}, p_{\delta})
    \label{dist_matching_proposed}
\vspace{-2mm}
\end{equation}

\noindent The function $l(\textbf{c}_x, \textbf{c}_y, \textbf{c}_z)$ maps trajectory coefficients to penalties on higher-order motion derivatives such as jerk. It also penalizes  violations of acceleration and velocity limits. The second term in \eqref{dist_matching_proposed} is the MMD-based distribution matching cost derived in the previous subsection. Please note $l_{dist}$ as presented in \eqref{mmd_ldist} is a function of $d_{ijk}$ which in turn is a function of the query point. In the context of trajectory optimization, the query points belong to the planned trajectory and are thus a function of $(\textbf{c}_x, \textbf{c}_y, \textbf{c}_z)$. In other words, different values of trajectory coefficients will map to distribution matching cost $l_{dist}$. 

We solve \eqref{dist_matching_proposed} through gradient-free CEM following closely the steps outlined in \cite{ccovoxel} and \cite{iCEM}. We provide only a brief overview here. We first draw different random values of $(\textbf{c}_x, \textbf{c}_y, \textbf{c}_z)$ from a Gaussian Distribution and evaluate the cost \eqref{dist_matching_proposed} on all of these generated samples. We then identify some best-performing samples and then fit a new Gaussian distribution on them for sampling in the subsequent iteration.

\subsection{SCP-MMD planner}

\noindent  A potential disadvantage of CEM-based trajectory optimization is that it requires a large number of cost function evaluations, which would translate to a large number of distance queries. Although our CEM planner can reach a peak re-planning of around $10Hz$, here we present an alternate trajectory optimizer with a lower computation time. To this end, we simplify the assumption that the obstacle size and orientation are deterministic. We obtain the nominal plane estimates (section ~\ref{nominal_plane}) and account for uncertainty by increasing the footprint of the nominal plane such that it bounds a large fraction (about 95\%) of the uncertain planes. 

\begin{subequations}\footnotesize
\begin{align}
    &\sum_t \ddot{x}(t)^2+\ddot{y}(t)^2+\ddot{z}(t)^2 \label{cost_scp}  \\
     & (x(t_0), y(t_0), z(t_0) ,\dot{x}(t_0), \dot{y}(t_0), \dot{z}(t_0), \ddot{x}(t_0), \ddot{y}(t_0) ,\ddot{z}(t_0) ) = \textbf{b}_0. \label{boundary_cond_initial}\\
    &(x(t_f), y(t_f), z(t_f) ,\dot{x}(t_f), \dot{y}(t_f), \dot{z}(t_f),  \ddot{x}(t_f), \ddot{y}(t_f), \ddot{z}(t_f) ) = \textbf{b}_f. \label{boundary_cond_final}\\
    &-v_{max} \leq \{\dot{x}(t), \dot{y}(t), \dot{z}(t)\}\leq v_{max}, \label{vel_bounds}  \\
    & -a_{max} \leq \{\ddot{x}(t), \ddot{y}(t), \ddot{z}(t)\}\leq a_{max}\label{acc_bounds} \\
    &d_c(x(t), y(t), z(t)) \geq 0, \forall t \label{sdf_scp}
\end{align}
\vspace{-3mm}
\end{subequations}
\normalsize



\noindent The cost function \eqref{cost_scp} minimizes the norm of the acceleration input over the planning horizon. The equality constraints \eqref{boundary_cond_initial}-\eqref{boundary_cond_final} enforces the boundary conditions on the trajectory. The inequalities \eqref{vel_bounds}-\eqref{acc_bounds} are the bounds on the velocities and accelerations. The last set of constraints \eqref{sdf_scp} ensure that the signed-distance at the query point $x(t), y(t), z(t)$ with respect to the bounding cuboid obstacle $c$ is greater than zero. To solve \eqref{cost_scp}-\eqref{sdf_scp}, we adopt a way-point parametrization of the trajectory and approximate the derivatives through finite differences.

\noindent \textbf{MMD based Initial Guess:} Optimization \eqref{cost_scp}-\eqref{sdf_scp} is non-convex due to the presence of \eqref{sdf_scp}. Thus, we need to provide an initial guess for its solution that will be used to kick-start an SCP-based optimizer. Typically, the initialization trajectory is computed through sampling-based planners. However, we take a fundamentally different approach. We draw a set of random trajectories from a Gaussian distribution centered around the straight line connecting the start and goal location and covariance given by \cite{STOMP}. We then compute the MMD distribution matching cost $l_{dist}$ over all these trajectories and choose one with the lowest value. This lowest cost trajectory will be used as the initial guess for solving \eqref{cost_scp}-\eqref{sdf_scp}.

\begin{remark}\label{scp_mmd}
Our SCP-MMD is equivalent to performing just one iteration of the CEM-based planner followed by refinement of the best trajectory through SCP. 
\end{remark}

\vspace{-1mm}

%% file: results.tex
In this section, we demonstrate that voxel-grid representations can be ineffective in a monocular setting due to the inherent sparsity and noise in the triangulated point clouds.  Furthermore, we show that  \textit{UrbanFly} can generate long-range smooth trajectories and execute them without the failure or drift in VI-SLAM.  Finally, we provide an empirical analysis of the convergence of CEM. 

\subsection{Simulation Setup}
\noindent We used Airsim \cite{airsim}, a state-of-the-art simulation framework which implements a physics engine, flight controller, and photorealistic scene for multirotor. Airsim runs inside Unreal Engine \cite{unrealengine} Editor. Our planners are implemented on a computer with 16GB RAM, core i7-10710 CPU. A dedicated simulation server equipped with NVIDIA-1070 graphics card, AMD Ryzen 7 3800x 8-core processor × 16 CPU, and 64GB RAM is used to run the Unreal Engine. The mapping module and CEM planner are programmed in C++. The SCP-MMD planner was implemented in Python using CVXPY\cite{cvxpy} for optimization.



\subsection{Simulation Environments} \label{Simulation_env_Details}
\noindent We test our approach on the following two data-sets.
\subsubsection{SquareStreet (Synthetic Dataset)}
Thanks to Unreal Engine (UE) Editor \cite{unrealengine}, we were able to create custom scenes to test our approach. We created a street in the shape of a square (Fig. 1(c)), having 47 buildings spread over 0.16 square kilometers. The environment is feature-rich and, therefore, can be easily used for benchmarking monocular vision-based perception and planning pipelines. The whole environment is modular and can be easily expanded or configured to meet custom requirements.


\subsubsection{Yingrenshi (Real city model)}
UrbanScene3D \cite{UrbanScene3D} is a dataset with real and virtual city models which can be imported into Unreal Engine. Yingrenshi, shown in \ref{fig:urbanscene_qual}. It has 252 buildings spread over 1sq. Km of area. This model is a replica of real-world buildings and streets. This environment helps us to evaluate our approach in a real-world setting.

\subsection{Baseline Comparison and Performance Analysis} \label{comparision}

\begin{figure}[!t]
    \begin{subfigure}[b]{0.45\linewidth}
    \centering
        \includegraphics[width=\linewidth]{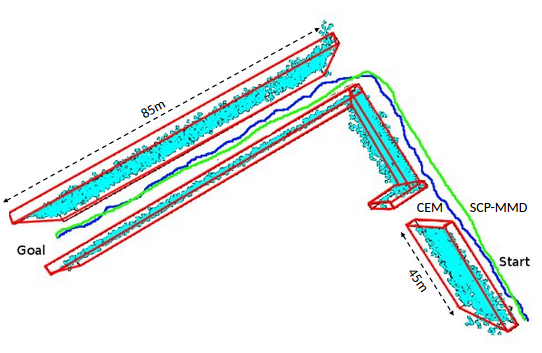}
        \subcaption{ \footnotesize Demonstration of obstacle reconstruction and trajectory execution of the the proposed planners in \textit{Yingrenshi} }
        \label{fig:urbanscene_global_map}
    \end{subfigure}
    \hfill
    \begin{subfigure}[b]{0.45\linewidth}
    \centering
        \includegraphics[width=0.45\textwidth, height=0.35\textwidth]{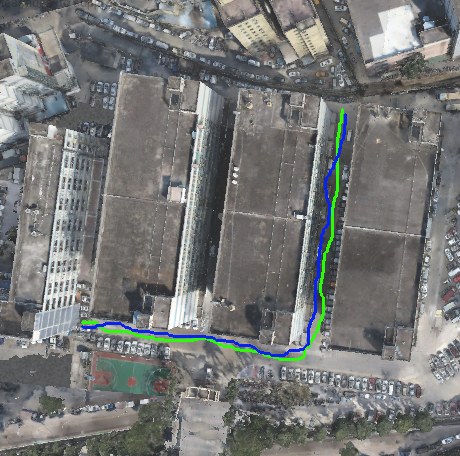}
        \includegraphics[width=0.45\textwidth, height=0.35\textwidth]{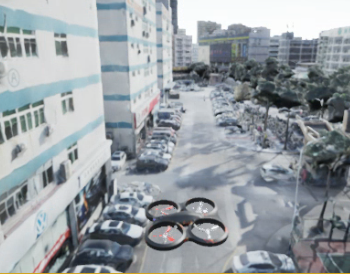}\\[1mm]
        \includegraphics[width=0.45\textwidth, height=0.35\textwidth]{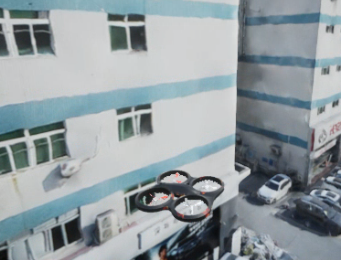}
        \includegraphics[width=0.45\textwidth, height=0.35\textwidth]{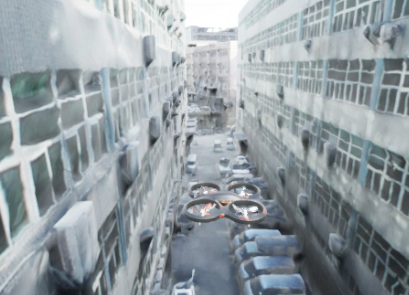}
        \subcaption{\footnotesize The visualization of the trajectories (top-left) in  \textit{Yingrenshi}}
        \label{fig:urbanscene_qual}
    \end{subfigure}
    \caption{ \footnotesize  Qualitative Results: \ref{fig:urbanscene_global_map} demonstrate the ability of \textit{UrbanFly} to reconstruct planar obstacle boundaries from triangulated point clouds and plan trajectories in real-time. In these figures, trajectories planned by SCP-MMD are depicted in Green color, and CEM trajectories are shown in Blue. In Fig. \ref{fig:urbanscene_qual}, the top-left tile shows the top view of the scene and the executed trajectory of 160m. Other tiles show three different viewpoints along the trajectory while avoiding large buildings.}
    \label{fig:qualitative}
    \vspace{-6mm}
\end{figure}

We compare the performance of \textit{UrbanFly} ( both CEM and SCP-MMD) with voxel grid based planners \cite{ccovoxel} \cite{fastplanner}. More than 100 trials were conducted in each simulation environment with various start and goal configurations.  

CCO-VOXEL \cite{ccovoxel} is a robust planning framework agnostic to the underlying uncertainty in the noisy distance map. FastPlanner \cite{fastplanner} is one of the state-of-the-art motion planning algorithms known to generate optimal and safe trajectories in noiseless environments. However,  both \cite{ccovoxel}, and \cite{fastplanner}  require dense point clouds to generate voxel maps, whereas the triangulated point cloud from a monocular VI-SLAM is sparse.

Both the \textit{UrbanFly} optimizers demonstrated superior performance over the competing baselines primarily due to two reasons. Firstly, the VI-SLAM pipeline would not faithfully reconstruct the featureless regions of the buildings, leading to an incorrect perception of the world. Specifically, the planner will falsely consider all the unoccupied regions as free and generate a trajectory through the walls.  Secondly, the noisy point clouds would lead to smearing of voxels; this leads to an increase in the clutter level, and the planner would fail, this situation is depicted in Fig.~\ref{teaser} \textcolor{red}{(d)}. On the other hand, our trajectory optimizers works with the polygonal representation of the world while accounting for the uncertainty in its estimation. Hence it can generate a safe and smooth trajectory.



Table ~\ref{table:benchmark} compares the performance of our trajectory optimizers with voxel grid-based planners \cite{ccovoxel}, \cite{fastplanner}, the latter being a deterministic planner that operates under the assumption of zero perception noise. For \textit{SquareStreet} environment, we observed that both our CEM and SCP-MMD planners were able to achieve more than \textbf{4} times improvement in success-rate over \cite{ccovoxel}, \cite{fastplanner}. The improvement in smoothness achieved by both stands at \textbf{39.34\%} and \textbf{42.43\%}. A similar trend can be observed for the \textit{Yingrenshi} environment in Table ~\ref{table:benchmark}. Both \textit{CEM} and \textit{SCP-MMD} planners outperform the competing baselines by \textbf{3.5} times in terms of success rate and demonstrated an improvement of close to \textbf{13\%} in terms of smoothness cost. 


We also evaluated our planners and the baselines regarding how far they enable the quadrotor to traverse before a failure of VI-SLAM. The results are summarized in the sixth and seventh columns of Table \ref{table:benchmark}. Both our trajectory optimizers achieved similar performance and were 8 times better than \cite{ccovoxel}, \cite{fastplanner}.

The computation time (5th column Table \ref{table:benchmark}) of our CEM-based trajectory optimizer is comparable with the closest baseline \cite{ccovoxel} that also performs uncertainty-aware planning. Both approaches used a similar number of distance queries, which were obtained using different methods. While our approach used analytical SDF \eqref{box_sdf}, \cite{ccovoxel} relied on distance queries from a voxel grid representation of the environment. Thus, our approach is expected to scale better if the problem complexity demands an increase in the sample size of CEM. Our SCP-MMD planner is faster than \cite{CCO_distance_to_collision} while unsuprisingly the deterministic planner \cite{fastplanner} has the lowest run-time. The qualitative demonstration of \textit{UrbanFly} in \textit{UrbanScene} is depicted in Fig ~\ref{fig:qualitative} and on \textit{SquareStreet} in Fig. 1(e) .




\begingroup
\renewcommand{\arraystretch}{1.6} 
\begin{table}[] 
\caption{Benchmark Comparison }
\centering  
\resizebox{0.45\textwidth}{!}{
\begin{tabular}{|l| l |r |r |r |r|r|}  

\hline                   
\hline
Environment & Method  
   & Smoothness & Success & Compute time & \multicolumn{2}{c|}{Traversed Length} \\\cline{6-7}
  & & ($m^2/s^5$) & \% & time (s)   & mean(m) & std(m)\\
\hline
\multirow{4}{1.5ex}{SquareStreet} & Ours (CEM)&  $\mathbf{8.82}$ & $\mathbf{86.666}$ & $0.095$  & $52.44$ & $12.37$ \\
&Ours (SCP-MMD) & $18.63$ & $82.51$ & $0.015$   & $51.35$ & $10.23$  \\
& CCO Voxel \cite{ccovoxel} & $ 14.54$ & $16.66$  & $0.097$ &$6.44$ & $1.37$ \\
& FastPlanner \cite{fastplanner} & $15.32$ & $9.66$ & $0.018$ &$9.44$ & $2.87$  \\
\hline                          
\multirow{4}{1.5ex}{Yingrenshi} & Ours (CEM)&  $\mathbf{10.82}$ & $\mathbf{83.87}$ & $0.103$   & $150.214$ & $ 16.304$ \\
&Ours (SCP-MMD) & $20.85$ & $80.47$ & $0.030$   & $133.82$ & $21.53$  \\
& CCO Voxel \cite{ccovoxel} & $ 12.54$ & $18.42$  & $ 0.089$& $46.44$ & $1.37$ \\
& FastPlanner \cite{fastplanner} & $13.54$ & $11.36$ & $0.016$ &$35.44$ & $4.67$  \\
\hline
\end{tabular}}
\label{table:benchmark}
\vspace{-4mm}
\end{table}
\endgroup

\vspace{-2mm}
\subsection{CEM Convergence Analysis} 
\noindent Here we empirically validate the ability of CEM to minimize the cost function and obtain an optimal trajectory. The convergence of CEM is shown in Fig.~\ref{CEM_optimization} where the cost profile of the mean trajectory has converged after $6$ iterations. Furthermore, a monotonic decrease in variance implies that with every iteration, the optimizer needs to progressively sample closer to the mean trajectory to find low-cost regions. 

Fig. ~\ref{CEM_optimization} depicts the convergence of distribution of constraint violation function (defined in ~\eqref{const_violation}) to the Dirac-Delta distribution (the ideal distribution as defined in Remark ~\ref{des_dirac}) over the CEM iterations. These convergence results have been empirically verified over large number of trials and validates the observations made in Remark \ref{dirac_sim}.
 
 
 \vspace{-2mm}
\subsection{Ablations}
 \vspace{-2mm}

\begin{figure}[!t]
\centering
\includegraphics[height=3cm]{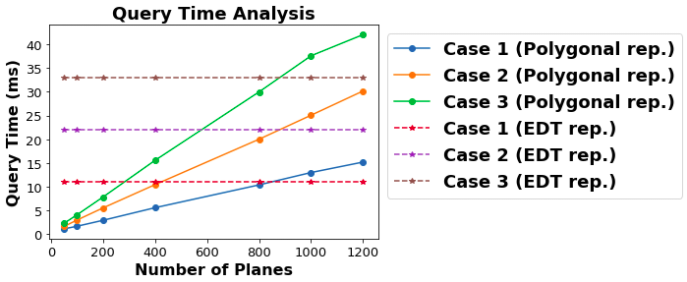}
	\caption{\footnotesize The query times recorded for both EDT and cuboid representation for multiple query points and multiple planes. In \textit{Case 1} we test with $50000$ query points, similarly for \textit{Case 2} and \textit{Case 3} we record the computation time for $100000$ and $150000$ query points respectively.  }
		\label{Query}
\end{figure}

\begin{figure}[!t]
\centering
\includegraphics[scale = 0.4]{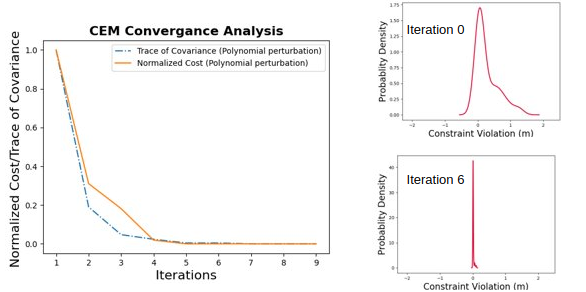}
	\caption{\footnotesize Left Figure: The normalized cost decreases and converges as iterations progress. With every iteration, we observe that the distribution of the collision-constraint violation approaches the Dirac delta function and therefore the MMD cost (\ref{mmd_ldist}) is minimized. Note that since plotting is based on approximate kernel density estimation from finite samples, a tiny part of the distribution appears to the left of 0 as well. }
		\label{CEM_optimization}
		\vspace{-7mm}
\end{figure}

\noindent Here we demonstrate a key advantage of the polygonal representation; fast distance queries.  The polygonal obstacle representation allows us to query an analytical function  ~\eqref{box_sdf}) to perform rapid collision checking. Fig. ~\ref{Query} depicts the results of the querying experiment. Our batch querying formulation provides a significant improvement in querying speed over the traditional \textit{Euclidian Distance Transform} (EDT) (\href{https://github.com/OctoMap/octomap/tree/devel/dynamicEDT3D}{dynamicEDT3D}) .
\vspace{-1mm}

%% file: conclusion.tex
In this paper, we demonstrated that constructing a polygonal representation of the world can significantly help to counter the inherent sparsity and uncertainty of the point clouds generated by the VI-SLAM system. We developed two trajectory optimizers that can leverage our chosen world model and obtain fast distance queries which was key in deriving an optimal and smooth trajectory in real-time. Both our optimizers showed improvement over several strong baselines in terms of success rate, trajectory smoothness, arc length, etc. In the future, we also aim to expand our approach to include both perception and quadrotor's state uncertainty. 


